\documentclass[11pt,onecolumn]{article}

\usepackage[margin=1in]{geometry}
\usepackage{amsmath,amssymb,amsthm,mathtools}
\usepackage{hyperref}
\usepackage{microtype}
\usepackage{booktabs}
\usepackage{array}
\usepackage{graphicx}
\usepackage{enumitem}

\title{Semantic Substrate Dynamics Theory:\\An Operator-Theoretic Framework for Geometric Semantic Drift}
\author{
Stephen Russell\\
Intelligent Systems and Robotics Department\\
University of West Florida\\
\texttt{russell@uwf.edu}
}
\date{}

\newtheorem{definition}{Definition}
\newtheorem{proposition}{Proposition}

\newcommand{\R}{\mathbb{R}}
\newcommand{\X}{\mathcal{X}}
\newcommand{\Wone}{\mathsf{W}_1}
\newcommand{\JS}{\operatorname{JS}}

\newcommand{\kappaOR}{\kappa_{\mathrm{OR}}}

\begin{document}
\maketitle

\begin{abstract}
Studies of semantic drift report heterogeneous signals, including embedding displacement, neighbor change, distributional divergence, and recursive trajectory instability, without a shared account that relates them. Semantic Substrate Dynamics Theory (SSDT) treats these signals as observables of one time-indexed substrate, $S_t=(X,d_t,P_t)$, that couples embedding geometry to a local diffusion kernel. The contribution is commensurability with a mechanism layer: the substrate separates within-basin churn from basin crossing, recursion-induced instability, and intervention-order effects, distinctions that a single detection score does not recover. Coarse Ricci curvature functions as a dense structural descriptor of basin and bridge geometry across the graph, and bridge mass, a node-level aggregate of incident negative curvature, functions as a sparse descriptor of the genuine bridge structure that is typically uncommon in embedding graphs. For recursive generation, node displacement relative to an origin decomposes into a radial component and a tangential component, which separates bounded departure from continuing reinterpretation. The predictions are stated in falsifiable form with a pre-declared rejection rule, and the predicted leading indicator of future rewiring is a local density statistic rather than the curvature aggregate. This manuscript provides the formal model, the assumptions, the observable roles, and the test contracts; empirical performance is deferred.
\end{abstract}

\section{Introduction}
Semantic drift is usually treated as a detection problem: something changed, therefore an alert should be raised. That orientation is operationally useful but weak on explanation, because the same drift score can arise from very different generative regimes, including benign local churn inside a stable semantic basin, boundary crossing between basins, or instability induced by an iterative transformation process. Without a mechanistic account the score is harder to trust, debug, and govern. A second difficulty is structural fragmentation. Separate lines of work track different quantities and label all of them drift, which leaves results hard to compare and lets interpretation itself drift after the fact.

The quantities prior studies generally track, embedding displacement, neighbor change, and trajectory stability among them, are each defensible for what they measure, so the aim here is not to replace them but to place them in one formal construct in which each has a defined role. Semantic Substrate Dynamics Theory (SSDT) names that construct the semantic substrate and develops its formalism together with pre-specified predictions. The substrate treats displacement, neighborhood rewiring, recursive instability, and intervention-order effects as observables of a single object, which supplies the mechanism layer that a detection score lacks and lets within-basin churn be told apart from basin crossing and from recursion-induced instability. The predictions are stated in falsifiable form with a rejection rule fixed in advance, so the framework is answerable to evidence rather than to conceptual appeal.

\section{Literature Background}
Diachronic distributional and embedding methods made large-scale semantic change measurable, especially across long historical corpora \cite{kutuzov2018survey,hamilton2016laws}. Standardized evaluation protocols then improved comparability for unsupervised lexical-change measurement \cite{arora2016randwalk,semeval2020}, and analyses of frequency and modeling artifacts showed that apparent laws of change can arise from representation and sampling effects unless controls are explicit \cite{dubossarsky2017outta}. This body of work established that a change in what a symbol means shows up as a change in its distributional neighborhood and not only as motion of its vector, which is the distinction between relational and translational drift.

Coarse Ricci curvature supplies an operator-level notion of local contraction for Markov processes on metric spaces \cite{ollivier2007,ollivier2009}. Graph adaptations relate curvature to local clustering and to bottleneck structure, which makes it a neighborhood-level diagnostic of where a graph is densely interconnected and where it narrows to a bridge \cite{jostliu2014}. Contextualized and dynamic embeddings raise the temporal resolution at which these quantities can be measured, though they leave the various mechanism labels loosely coupled to one another \cite{bamlermandt2017,giulianelli2020}. A separate regime arises when generation is recursive, so that the output of one step becomes the input to the next, as in iterative rewriting, summarization, and multi-turn dialogue. Drift there is a property of a trajectory rather than a comparison of two snapshots, and stability, basin structure, and persistence become the relevant descriptors. Across all of these settings the individual measurements are understood well enough in isolation, and what remains unsettled is how they relate. The gap is not metric availability; what is missing is a unified theoretical model that standardizes the definitions of node drift, edge fragility, and recursive instability, while clearly distinguishing their individual mechanics.

\section{Semantic Substrate}
\begin{definition}[Semantic substrate]
For each analysis window $t$, the semantic substrate is defined as $S_t=(X,d_t,P_t)$, where $\X$ is a set of semantic objects, $d_t$ is an embedding-induced metric on $\X$, and $P_t$ is a one-step Markov diffusion kernel.
\end{definition}

The metric $d_t$ is instantiated through an embedding map $f_t:X\to\R^{d}$, from which a neighborhood graph $G_t=(V,E_t)$ on $V=X$ is built by kNN or mutual-kNN. For each node $x$, the local neighborhood measure $m_x^{(t)}$ is

\begin{equation}
 m_x^{(t)}(z)=
 \begin{cases}
 \alpha, & z=x,\\[2pt]
 (1-\alpha)\,\dfrac{w_{xz}}{\sum_{u\in N_t(x)}w_{xu}}, & z\in N_t(x),\\[6pt]
 0, & \text{otherwise},
 \end{cases}
 \label{eq:local_measure}
\end{equation}
with idleness $\alpha\in[0,1]$ and edge weights $w_{xz}>0$, and the diffusion kernel is set to $P_t(x,\cdot)=m_x^{(t)}(\cdot)$. Equation \eqref{eq:local_measure} exposes the two construction knobs that recur throughout the framework: the idleness $\alpha$, which fixes how much mass a node retains on itself, and the weighting rule $w_{xz}$, which fixes how the remaining mass spreads over neighbors. Their effect on curvature is treated in Section \ref{sec:curvature}. Drift occurs when geometry ($d_t$), diffusion ($P_t$), or both evolve across windows.

\section{Four Drift Modes as Observables of One Substrate}
\label{sec:modes}
The substrate decomposes drift into four non-equivalent modes, summarized in Table \ref{tab:modes}. Each mode is an observable of the same object, which is what lets a single score be resolved into a mechanism rather than reported as an undifferentiated alarm.

\begin{table}[ht]
\centering
\caption{The four drift modes and the substrate observable that defines each.}
\label{tab:modes}
\begin{tabular}{@{}l>{$}l<{$}p{7.1cm}@{}}
\toprule
Mode & \textrm{Observable} & What it measures \\
\midrule
Translational & D_{\mathrm{tr}}(x;t_0,t_1) & displacement of a node's own embedding \\
Rewiring & D_{\mathrm{rw}}(x;t_0,t_1)=\JS(m_x^{(t_0)}\Vert m_x^{(t_1)}) & redistribution of a node's neighborhood \\
Dynamical & x_{k+1}=F(x_k) & trajectory behavior under an iterated operator \\
Process & H_t=G_t\circ F_t & change from composing evolution with intervention \\
\bottomrule
\end{tabular}
\end{table}

\subsection{Translational drift}
A direct displacement observable is
\begin{equation}
D_{\mathrm{tr}}(x;t_0,t_1)=\|f_{t_1}(x)-f_{t_0}(x)\|,
\end{equation}
with cosine-based analogs as alternatives. This mode captures motion in embedding space and is vulnerable to global embedding reconfiguration and alignment effects, so a large value can reflect a rotation of the whole space rather than a change in what the node means.

\subsection{Rewiring drift}
Neighborhood-distribution shift at node $x$ is
\begin{equation}
D_{\mathrm{rw}}(x;t_0,t_1)=\JS\!\left(m_x^{(t_0)}\,\Vert\,m_x^{(t_1)}\right),
\label{eq:rewire}
\end{equation}
with entropy change and transport shift as companions,
\begin{equation}
\Delta H(x)=\left|H\!\left(m_x^{(t_1)}\right)-H\!\left(m_x^{(t_0)}\right)\right|,
\qquad
\Delta W(x)=\Wone\!\left(m_x^{(t_0)},m_x^{(t_1)}\right).
\end{equation}
These quantities measure redistribution of the local neighborhood rather than global translation, so they register when a node keeps its position but reorganizes the company it keeps, which is the case that translational drift misses. Rewiring drift is the outcome that the theory's predictions target.

\subsection{Dynamical drift}
\label{sec:dynamical}
Let an iterative semantic operator be
\begin{equation}
x_{k+1}=F(x_k),
\end{equation}
where $F$ represents recursive rewriting, summarization, or transformation followed by projection into $\X$. Drift is then a trajectory property, including sensitivity, basin switching, and persistence, and not only a two-window difference.

Trajectory geometry admits a decomposition that separates how far a state has moved from which direction it has taken. Fix an origin $x_0$, the seed of the recursion, and write the radial distance
\begin{equation}
r_k = d(x_k, x_0).
\end{equation}
The one-step displacement $\varepsilon_k = x_{k+1}-x_k$ splits into a radial part along the outward direction $u_k = (x_k-x_0)/\lVert x_k-x_0\rVert$ and a tangential part orthogonal to it,
\begin{equation}
\varepsilon_k = \varepsilon_k^{\parallel} + \varepsilon_k^{\perp},
\qquad
\varepsilon_k^{\parallel}=\langle \varepsilon_k, u_k\rangle\,u_k,
\qquad
\varepsilon_k^{\perp}=\varepsilon_k-\varepsilon_k^{\parallel},
\end{equation}
so that radial velocity is $r_{k+1}-r_k$ and tangential velocity is $\lVert\varepsilon_k^{\perp}\rVert$. The two components describe different channels. Radial departure tends to be self-limiting, because under a bounded metric such as cosine distance the distance from an origin cannot grow without limit and a trajectory settles toward a characteristic radius. Tangential motion is not bounded in the same way, so a trajectory can continue to change direction at approximately fixed distance from its origin. Two trajectories seeded from the same $x_0$ can therefore remain near the same radius $r_k$ while their mutual distance $d(x_k^{(i)},x_k^{(j)})$ grows, so the origin constrains how far interpretation moves without fixing which interpretation it settles into. This decomposition is descriptive geometry rather than a forecasting device.

\subsection{Process drift (composition of evolution and intervention)}
Observed drift in deployed systems is generated jointly by semantic evolution and pipeline interventions. Let $F_t$ denote organic semantic evolution, for example corpus shift or endogenous updates, and let $G_t$ denote intervention, for example retrieval grounding, policy filtering, safety tuning, or decoding constraints. The composed update operator is
\[
H_t = G_t \circ F_t,
\]
and process drift is the change this composition introduces beyond either factor in isolation, which separates what the semantics would have done from what the pipeline made them do. The composition also implies order effects when the two do not commute,
\[
G_t \circ F_t \neq F_t \circ G_t,
\]
so applying an intervention before rather than after semantic evolution can produce different neighborhoods, curvature profiles, and rewiring risk even when the same components are used.

\section{Curvature and Structural Description}
\label{sec:curvature}
\begin{definition}[Coarse Ricci curvature]
For $x\neq y$ with $d_t(x,y)>0$,
\begin{equation}
\kappaOR^{(t)}(x,y)=1-\frac{\Wone\!\left(m_x^{(t)},m_y^{(t)}\right)}{d_t(x,y)}.
\label{eq:orc}
\end{equation}
\end{definition}

Positive curvature corresponds to locally contractive one-step diffusion and negative curvature to locally expansive, bridge-like structure. The contractivity result imported from the operator theory ties the local sign to a global stability statement.

\begin{proposition}[Contractivity under positive curvature]
The following is a standard consequence of coarse Ricci curvature \cite{ollivier2007,ollivier2009}.
Let $\kappa_0=\inf_{x\neq y}\kappaOR^{(t)}(x,y)$. If $\kappa_0>0$, then for all probability measures $\mu,\nu$ on $\X$,
\begin{equation}
\Wone(\mu P_t,\nu P_t)\le (1-\kappa_0)\,\Wone(\mu,\nu),
\label{eq:contractivity}
\end{equation}
and consequently
\begin{equation}
\Wone(\mu P_t^k,\nu P_t^k)\le (1-\kappa_0)^k\,\Wone(\mu,\nu).
\end{equation}
\end{proposition}

Equation \eqref{eq:contractivity} is an operator-level stability condition that complements node-level drift magnitudes. The hypothesis $\kappa_0>0$ is a local, basin-restricted condition rather than a global one, since curvature is negative on the bridge structure the framework targets, so $\kappa_0$ is generically at or below zero across the whole graph and the bound characterizes basin interiors rather than the entire substrate.

Curvature is a dense structural descriptor. It is defined at every edge, applies across the whole graph, and labels each neighborhood as basin-like or bridge-like, which localizes where the semantic topology is easiest to reconfigure and gives the mechanism layer that a detection score lacks. Its role in the framework is characterization and, in applied settings, complementary detection, not by itself the population-level forecasting of rewiring. Node-level bridge mass aggregates incident negative curvature,
\begin{equation}
B_t(x)=\sum_{y\in N_t(x)}\pi_t(x,y)\,\bigl(-\kappaOR^{(t)}(x,y)\bigr)_+,
\label{eq:bridge_mass}
\end{equation}
where $(u)_+=\max(u,0)$, $\pi_t(x,y)\ge 0$, and $\sum_{y\in N_t(x)}\pi_t(x,y)=1$. A large $B_t(x)$ indicates that a substantial share of incident neighborhood structure lies in the bridge-like regime. Bridge mass is a sparse descriptor: it identifies the specific nodes that sit on genuine bottlenecks, but genuine bridge structure is uncommon in typical embedding graphs, so the quantity is near zero across the bulk of a vocabulary and takes appreciable values only in a thin tail. That sparsity is the reason bridge mass belongs to the descriptive role, as a marker of true bridge nodes, rather than to the population-level forecasting role. Table \ref{tab:roles} records the role assigned to each substrate quantity.

\begin{table}[ht]
\centering
\caption{Role assigned to each substrate quantity, separating description from prediction.}
\label{tab:roles}
\renewcommand{\arraystretch}{1.25}
\begin{tabular}{@{} >{\raggedright\arraybackslash}p{4.9cm} >{\raggedright\arraybackslash}p{4.5cm} >{\raggedright\arraybackslash}p{5.0cm} @{}}
\hline
Quantity & Role & Scope \\
\hline
Coarse Ricci curvature $\kappaOR$ & dense structural descriptor & basin versus bridge geometry across the whole graph \\
\hline
Bridge mass $B_t$ & sparse structural descriptor & genuine bridge nodes, a thin tail of the distribution \\
\hline
Self-neighbor similarity $C_k$ & predicted leading indicator of rewiring & population level \\
\hline
Radial component $r_k$ & bounded-departure descriptor & recursive generation \\
\hline
Tangential component $\varepsilon_k^{\perp}$ & persistent-reinterpretation descriptor & recursive generation \\
\hline
Rewiring drift $D_{\mathrm{rw}}$ & outcome and criterion & the quantity predictions target \\
\hline
\end{tabular}
\end{table}

\begin{figure}[ht]
\centering
\includegraphics[width=0.8\linewidth]{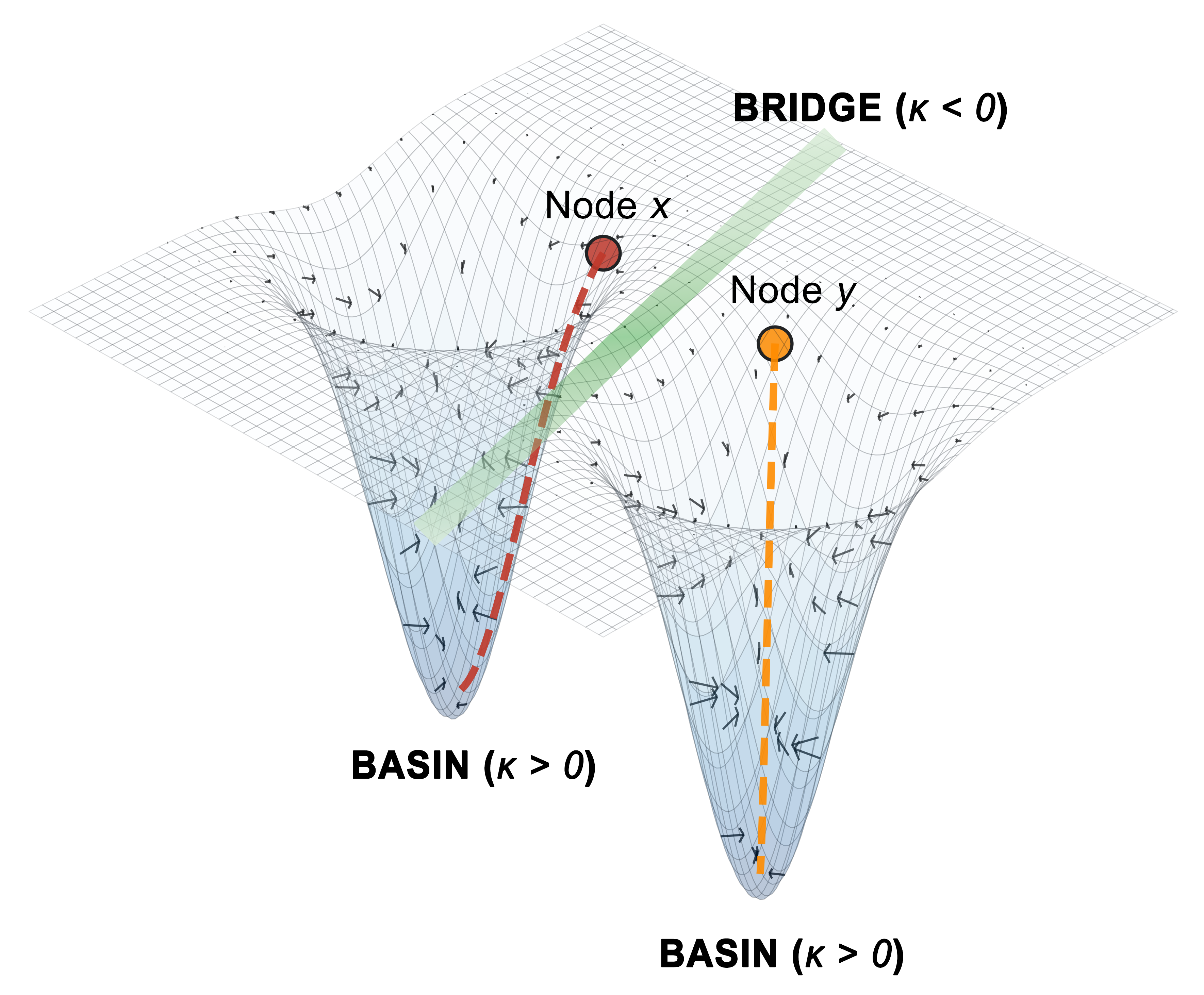}
\caption{\textbf{Semantic basins and bridges in substrate geometry.}
Blue wells depict positively curved basin regions ($\kappa>0$) where local diffusion tends to be contractive and trajectories are pulled toward stable neighborhoods. The green ridge indicates a negatively curved bridge regime ($\kappa\le 0$), where diffusion is locally expansive and connectivity is fragile. Nodes near or traversing bridge structure are more susceptible to neighborhood rewiring under perturbation or intervention.}
\label{fig:semantic_bridges_basins}
\end{figure}

Figure \ref{fig:semantic_bridges_basins} gives the visual intuition: positively curved basins are locally self-reinforcing, while negatively curved bridge regions are the high-leverage transition structure where small updates can redirect flow across neighborhoods. The figure also makes the sparsity concrete, since the bridge occupies a narrow ridge between two wide basins.

\subsection{Idleness and edge weighting}
\label{sec:alpha}
The curvature in Equation \eqref{eq:orc} is computed from the neighborhood measures, so it inherits the two knobs in Equation \eqref{eq:local_measure}. Writing the lazy measure as a mixture,
\begin{equation}
m_x^{(\alpha)}=\alpha\,\delta_x+(1-\alpha)\,\hat m_x,
\qquad
\hat m_x(z)=\frac{w_{xz}}{\sum_{u\in N_t(x)}w_{xu}}\ \text{for } z\in N_t(x),
\label{eq:lazy_mixture}
\end{equation}
makes the two effects explicit. The idleness $\alpha$ sets the self-retained mass. Larger $\alpha$ holds more mass at the node, which reduces the transport distance $\Wone(m_x^{(\alpha)},m_y^{(\alpha)})$ between adjacent lazy measures relative to $d_t(x,y)$ and therefore shifts $\kappaOR$ upward toward the contractive regime. The idleness thus sets the overall curvature scale, and the half-lazy choice $\alpha=0.5$ is a common reference point that keeps the walk from collapsing onto the node while retaining sensitivity to neighborhood shape. The weighting rule $w_{xz}$ determines how the non-idle mass distributes over neighbors, whether uniformly, by cosine similarity, or by a distance-decay kernel, and therefore determines which incident edges register as negative curvature. Because both knobs move numerical curvature values, the quantity that should be reported and defended is the ordering of nodes rather than the absolute level, which is the content of assumption A5. The lazy construction produces its largest corrections at low-degree boundaries, where few neighbors are available to absorb transported mass, so boundary nodes are where the choice of $\alpha$ matters most and where stability over the $(\alpha,k,\text{weighting})$ grid should be checked.

\subsection{Computing the transport distance}
Coarse Ricci curvature requires a Wasserstein distance for every edge, and exact optimal transport is expensive when computed across many local comparisons. Entropic regularization yields the Sinkhorn objective
\begin{equation}
\Wone^{\varepsilon}(\mu,\nu)=\min_{\gamma\in\Pi(\mu,\nu)}\langle C,\gamma\rangle-\varepsilon H(\gamma),
\end{equation}
solved by matrix-scaling iterations and typically far cheaper in practice than exact linear programming \cite{cuturi2013}. Practical use specifies $\varepsilon$, the iteration budget, and the stopping tolerance, and can assess approximation error as a function of $\varepsilon$ and compute budget by cross-validating a subset of edges against exact optimal transport.

\section{Assumptions}
The theory depends on six assumptions that can be violated.

\begin{enumerate}[label=\textbf{A\arabic*:},leftmargin=2.2em]
\item \textbf{Local geometric validity:} The metric $d_t$ is locally meaningful for the domain, even if global geometry is imperfect.
\item \textbf{Neighborhood as conditional proxy:} The measure $m_x^{(t)}$ is an adequate approximation to local contextual meaning around $x$.
\item \textbf{Within-window stationarity:} The substrate is approximately stationary inside each analysis window.
\item \textbf{Confound control:} Frequency, utilization, and sampling effects are controlled or modeled explicitly.
\item \textbf{Graph robustness:} Qualitative conclusions persist under reasonable variation of $k$, $\alpha$, and weighting rules.
\item \textbf{Operator relevance:} For recursive studies, the chosen $F$ approximates an operational transformation of interest.
\end{enumerate}

\section{Testable Predictions and Falsification Criteria}
\label{sec:predictions}
The theory is evaluated through pre-specified predictions under a single rejection rule. The predictions separate a structural claim at the population level from dynamical and intervention claims within the recursive regime.

\begin{enumerate}[label=P\arabic*., ref=P\arabic*]
\item \textbf{Structural fragility and rewiring:} Nodes in loosely connected, low-density neighborhoods rewire more, and the leading indicator that survives frequency and sampling controls out of sample is a local density statistic, self-neighbor similarity
\begin{equation}
C_k(x)=\frac{1}{k}\sum_{y\in N_k(x)}\cos\!\bigl(f_t(x),f_t(y)\bigr),
\label{eq:selfsim}
\end{equation}
the mean similarity of a node to its $k$ nearest neighbors, with lower $C_k(x)$ associated with higher future rewiring $D_{\mathrm{rw}}(x;t,t+\Delta)$. The population-level forecasting role is assigned to $C_k$ rather than to the curvature aggregate $B_t$.
\item \textbf{Boundary concentration:} In its non-circular leading form, large rewiring events concentrate at nodes whose incident curvature shifts toward the negative regime in advance of the event.
\item \textbf{Intervention directionality:} Stabilizing interventions increase local contractivity and reduce future rewiring risk, and destabilizing interventions show the opposite pattern.
\item \textbf{Dynamics-geometry alignment:} Within closed recursive generation, trajectories that traverse negative-curvature regions have higher basin-switch rates and longer perturbation persistence.
\item \textbf{Non-commutativity effect:} Intervention order is not neutral, defined through
\[
\Delta_{\mathrm{comm}}(x,t)=d\!\left((G_t\!\circ\!F_t)(x),(F_t\!\circ\!G_t)(x)\right),
\]
where larger $\Delta_{\mathrm{comm}}(x,t)$ is associated with larger differences in endpoint stability and rewiring risk across intervention timing schedules.
\end{enumerate}

\paragraph{Minimal operationalization of P5.}
Fix two intervention families $G_t^{(1)}$, $G_t^{(2)}$ and one evolution family $F_t$ before inspecting outcomes. For each node $x$ and window $t$, evaluate matched schedules $G_t^{(i)}\!\circ\!F_t$ and $F_t\!\circ\!G_t^{(i)}$, $i\in\{1,2\}$, under identical budgets and confound controls. The endpoints, fixed before outcome inspection, are future rewiring drift $D_{\mathrm{rw}}(x;t,t+\Delta)$, basin-switch rate under recursive probing, and calibration error of rewiring-risk forecasts. The strict null is order neutrality: after controls, $\Delta_{\mathrm{comm}}(x,t)$ adds no incremental predictive value beyond baseline drift and frequency covariates.

\paragraph{Falsification rule.}
A prediction is rejected if it does not improve out-of-sample discrimination or calibration relative to controls and simpler baselines.

\section{Conclusion}
SSDT formulates semantic drift on one evolving substrate that renders translational movement, neighborhood rewiring, recursive instability, and intervention-order effects comparable within a single operator-geometric model. The central contribution is a decomposition in which multiple drift signals are coupled measurements of a shared latent evolution, which enables mechanistic distinctions between within-basin churn, basin transitions, and recursion-induced instability that detection-only scores do not recover, and which clarifies major failure modes including embedding anisotropy, graph instability, and window aliasing. Within that construct the curvature quantities function as structural descriptors, dense at the edge level in $\kappaOR$ and sparse in the bridge aggregate $B_t$, while the predicted leading indicator of future rewiring is a local density statistic rather than the curvature aggregate. The test contract of explicit assumptions, operational predictions, and a rejection rule defines the empirical program by which the framework is to be evaluated, through out-of-sample discrimination and calibration under confound controls and robustness analysis rather than through conceptual appeal.

\end{document}